\documentclass[5p]{elsarticle}
\usepackage{lineno} 
\modulolinenumbers[5]
\usepackage{times}
\usepackage{helvet}
\usepackage{courier} 
\usepackage{graphicx}
 
\usepackage{booktabs}    
\usepackage{amsfonts}    
\usepackage{nicefrac}    
\usepackage{microtype}   
\usepackage{amsmath}
\usepackage{algorithm}
\usepackage{algorithmic}
\usepackage{color}
\usepackage{lipsum} 
\usepackage{xargs}
\usepackage{subcaption}

\begin{document}
\begin{frontmatter}
\title{Graph Routing between Capsules}

\author[NPU]{Yang Li}
\ead{liyangnpu@nwpu.edu.cn}
\author[TUD]{Wei Zhao}
\ead{zhao@aiphes.tu-darmstadt.de}
\author[NTU]{Erik Cambria\corref{mycorrespondingauthor}}
\author[PSU]{Suhang Wang} 
\ead{szw494@psu.edu}
 \author[TUD]{Steffen Eger}
 \ead{eger@aiphes.tu-darmstadt.de}

\cortext[mycorrespondingauthor]{Corresponding author}
\ead{cambria@ntu.edu.sg}

\address[NPU]{Northwestern Polytechnical University, China}
\address[NTU]{Nanyang Technological University, Singapore} 
\address[TUD]{Technical University of Darmstadt, Germany}
\address[PSU]{Pennsylvania State University, USA}

\linenumbers
\begin{abstract}
 
Routing methods in capsule networks often learn a hierarchical relationship for capsules in successive layers, but the intra-relation between capsules in the same layer is less studied, while this intra-relation is a key factor for the semantic understanding in text data. Therefore, in this paper, we introduce a new capsule network with graph routing to learn both relationships, where capsules in each layer are treated as the nodes of a graph. We investigate strategies to yield adjacency and degree matrix with three different distances from a layer of capsules, and propose the graph routing mechanism between those capsules.
We validate our approach on five text classification datasets, and our findings suggest that the approach combining bottom-up routing and top-down attention performs the best. 
Such an approach demonstrates generalization capability across datasets. 
Compared to the state-of-the-art routing methods, the improvements in accuracy in the five datasets we used were 0.82, 0.39, 0.07, 1.01, and 0.02, respectively.

\end{abstract}
\end{frontmatter}

\section{Introduction}

Since capsule networks were proposed, they have been successfully applied in image processing~\cite{hinton2011transforming,sabour2017dynamic}. In recent years, a lot of works have appeared that also adopted capsule networks for text classification. For example, Zhao et al.~\cite{zhao2018investigating} investigate the effectiveness of capsule networks in small data settings, while Ren et al.~\cite{ren2018compositional} introduce a variant of capsule networks with deep compositional code learning in text classification. 
The success of capsule networks mainly relies on a bottom-up mechanism, namely routing by agreement~\cite{sabour2018matrix,sabour2017dynamic}, which routes low-level capsules to high-level capsules in order to learn hierarchical relationships between layers. This mechanism helps capsule networks capture spatial features well.

However, there are problems in applying capsule networks to text data. Generally, these routing algorithms do not account for the intra-relationships between capsules in each layer, while these intra-relationships exist in text data usually.
For example, in the sentence ``The battery has a long life'' with a positive sentiment, there is a part-whole relationship between the words ``battery'' and ``life'', a modification relationship between the words ``long'' and ``life'' and etc. 
These intra-relationships, often ignored by previous routing methods, help understanding the positive sentiment. 
It is imperative to leverage implicit knowledge in the words' semantic context to learn hierarchical and intra-relationships spontaneously across layers~\cite{camnt6,chimod,akthow}.

In this paper, we study the novel problem of simultaneously exploring these two relationships (intra-relationship and hierarchical relationship) interactively in the routing context. In essence, we need to solve several challenges: i) how to evaluate the relationship between capsules in the same layer; ii) how to further improve the quality of the learned features in text classification; and iii) how to evaluate the quality of the relationships that learn from the capsules.
Graph convolutional networks (GCN)~\cite{kipf2016semi} have demonstrated good effects in the graph feature learning~\cite{wu2020comprehensive,pan2019learning}.
And we believe this method also takes effect in the capsules feature gathering.
In an attempt to solve these challenges, we treat the capsule as a node in a graph and propose a new graph routing mechanism that learns the intra-relationships with GCN. 
In general, our proposed graph routing learns the intra-relationship by aggregating information from other capsules in the same layers, and learns the hierarchical relationship with routing by agreement across different layers. 
Thus, the differences between proposed graph routing and existing routing are:
\begin{itemize}

\item Our graph routing method pays attention to the intra-relationship learning with graph neural network, while existing routing methods do not;
\item Our graph routing method applies the attention mechanism in the feature strength, and existing routing methods do not;
\end{itemize}
In this paper, the relationship between different capsules is evaluated by the Wasserstein distance (WD) effectively, and a new normalization trick is proposed to approximate the adjacency matrix well. Also, implicit information in text data is identified easily with information aggregation and routing agreement. 
Compared with state-of-the-art models, the performance in three of the five datasets improved by more than 0.3 points in accuracy.
Therefore, the challenges mentioned before are solved well with the proposed graph routing, and the main contributions of this paper are:
\begin{itemize}
\item We investigate, for the first time, intra-relationships between capsules by leveraging words' semantic context where we treat capsules in each layer as nodes in a graph.
\item We introduce a new routing algorithm combining bottom-up routing and top-down attention and learn hierarchical and intra-relationships spontaneously. 
\item Extensive experiments show that our proposed routing algorithm performs better than existing routing methods.
\end{itemize}

\section{Preliminary}
\label{sec:preliminary}
When capsule networks were proposed~\cite{hinton2011transforming}, they were mainly applied in image processing. 
Multiple capsules were required to be consistent in one detection.
However, the intra-relationship between capsules cannot be ignored when processing text data. To our best knowledge, there are no works about intra-relationship learning in the routing. Before introducing our graph routing, we will make a preliminary introduction about the classic routing. 
The symbols that this paper uses are listed in Table~\ref{tab:sym}.
\begin{table*}[]
\centering
\begin{tabular}{l|p{4.3cm}|l|p{4.3cm}}
\hline
$\mathbf{x}$ & The representation of a document &$I_{N}$ & The identity matrix \\\hline
$\mathbf{u_{i}}$  & The first layer capsule vector in the routing & $\mathbf{s_{j}}$& The second layer capsule vector in the routing \\\hline
$\mathbf{v_{j}}$ & The output from the routing & $c_{ij}$ & The weight variable \\\hline
$W^{D}$  & The affine transformation matrix & $e_{ij}$ & The coupling coefficient \\\hline
$W_{i}$ & The parameters of the neural networks & $A,\tilde{A}$ & The adjacency matrix \\\hline
$D$ & The degree matrix in the graph & $f_{att}(\cdot)$ & The feed-forward function \\\hline
$d^{w}, d^{e},d^{c}$ & The Wasserstein distance, euclidean distance and cosine distance & $g^{a}_{i}$ & The output from N-gram convolutional layer \\\hline
$h_{j}$ & The output from GCN &$m_{j}$ & The output from $f_{att}$ \\\hline
$W^{a}, W^{b}$ & The filter in the convolutional layer & $p_{i}$ & The output from primary capsule layer\\\hline
$K$ & The size of N-gram & $s$ & The stride size \\\hline
$b_{1}, b_{2}$ & The bias term & $B_{1}, B_{2}$ & The number of the filter \\\hline
$L$ & The document length & $d$ & The capsule dimension \\\hline 
$E$ & The capsule number after the routing &$f_{1}(\cdot)$ & The activate function (i.e. ReLU) \\\hline 
 $C$& The class number& $f_{2}(\cdot)$ & The squash function \\\hline
$\alpha_{j}$ & The attention value & \\\hline
\end{tabular}
\caption{The list of symbols involved in this paper. }
\label{tab:sym}
\end{table*}

Since dynamic routing was proposed by~\cite{sabour2017dynamic}, it has been treated as the standard routing method. Let the capsule vector in the first layer be $\mathbf{u}_{i}$.
Before routing, a transformation procedure is applied to encode spatial relationships between local features and global features: 
\begin{equation*}
\hat{u}_{j|i}=W^{D}\mathbf{u}_{i} 
\end{equation*}
where $W^{D}$ is an affine transformation matrix.
Generally, routing decides how to send the capsule vector $\mathbf{u}_{i}$ to the second layer.
This is controlled by a weight variable $c_{ij}$, which is multiplied with the corresponding value $\hat{u}_{j|i}$: 
\begin{equation*}
\mathbf{s}_{j} = \sum_{i=1}^{n}c_{ij}\hat{u}_{j|i}
\end{equation*}
The squash function is then applied to ensure that the norm of $\mathbf{s}_j$ is bounded by 1, while the direction of $\mathbf{s}_j$ should not be changed. 
The squash function has the form 
\begin{equation*}
\mathbf{v}_{j} = \frac{||\mathbf{s}_{j}||^{2}}{1+||\mathbf{s}_{j}||^{2}}\frac{\mathbf{s}_{j}}{||\mathbf{s}_{j}||}
\end{equation*}
Generally, $c_{ij}$ is a non-negative scalar, and the sum of all weights $c_{ij}$ in the first layer is $1$.
It is obtained from the softmax function: 
\begin{equation*}
c_{ij} = \frac{\exp(e_{ij})}{\sum_{k}\exp(e_{ik})}
\end{equation*}
Different from attention models, routing is a bottom up method in feature gathering, and this 
is illustrated in following equation which is the across layer operation with $i$ being the capsules from the first layer and $j$ being the capsules from the second layer: 
\begin{equation*}
e_{ij} = e_{ij}+\hat{u}_{j|i}\mathbf{v}_{j}
\end{equation*}

$e_{ij}$ is the coupling coefficient, and the general routing method is using the clustering method to find the probability in mapping different capsules to related capsules across layers. 
However, as opposed to related works, we investigate relationships between capsules in the same layer. 

\section{Model Architecture}
\label{sec:architecture}
\begin{figure*}[ht]
 \centering
   \includegraphics[width=1\linewidth]{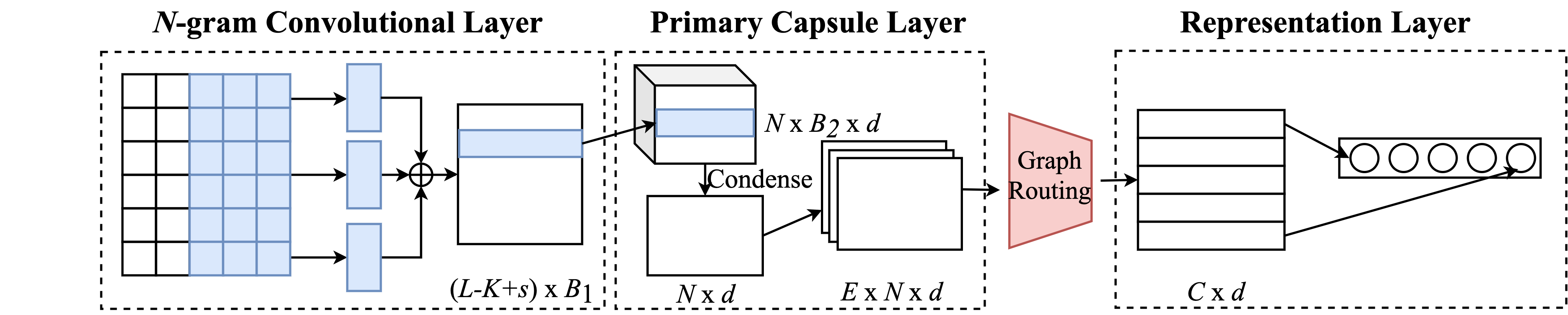} 
  \caption{The architecture of the model.}
  \label{fig:architecture}
 \end{figure*}

To make a fair comparison with other routing mechanisms, we are using the same architecture proposed by Zhao et al.~\cite{zhao2018investigating}.
The model architecture components are four layers which are the $N$-gram convolutional layer (NCL), the primary capsule layer (PCL), the routing layer (RL), and the representation layer. The architecture is basically the same as Capsule-A in~\cite{zhao2018investigating}, which is depicted in Figure~\ref{fig:architecture}.
 
\subsection{N-gram Convolutional Layer (NCL)}
After getting the input from the word embedding, each document is represented by $\mathbf{x}\in R^{L\times D}$, where $L$ denotes the document length (padded where necessary), $D$ is the word embedding length.
Then the input will be fed into the NCL to detect features at different positions.
The process is depicted in equation~\ref{equ:n-gram_cl},
\begin{equation}
\label{equ:n-gram_cl}
g_{i}^{a} = f_{1}(W^{a}\cdot \mathbf{x}_{i:i+K-s}+b_{1})
\end{equation}
where $K$ denotes the $N$-gram size, $s$ is the stride width, $W^{a}\in R^{K\times D}$ is the filter during the feature detection with convolutional operation, $b_{1}\in R$ is a bias term, and $f_{1}$ denotes the activation function (i.e., ReLU) in that layer.
After the filtering with a certain filter $W^{a}$ over the $\mathbf{x}_{i:i+K-s}$, we get a feature value $g_{i}^{a}\in R$. 
All of those features form the feature map $\mathbf{g}^{a}=[g_{1},g_{2},...,g_{L-K+s}]$ with $\mathbf{g}^{a}\in R^{L-K+s}$ from the NCL.
The number of the filter is set to $B_{1}$.
Hence, the output of the NCL is $\mathbf{g}$ which is rearranged as 
\begin{equation*}
\mathbf{g} = [\mathbf{g}_{1}, \mathbf{g}_{2},...,\mathbf{g}_{B_{1}}]\in R^{(L-K+s)\times B_{1}}
\end{equation*}
\subsection{Primary Capsule Layer (PCL)}
Then $\mathbf{g}$ is fed into the PCL which is composed of the convolutional operation.
In this layer, each row in $\mathbf{g}$ convolves with a filter $W^{b}\in R^{B_{1}\times d}$, where $d$ denotes the capsule dimension. 
This is depicted in equation~\ref{equ:pcl}.

\begin{equation}
\label{equ:pcl}
p_{i}=f_{2}(W^{b}\cdot \mathbf{g}_{i}+b_{2})
\end{equation}
where $f_{2}$ is the squash function for the output vector, $b_{2}$ is a capsule bias term.
 As in the NCL, there are $B_{2}$ filters in total.
 Therefore, the generated features are rearranged as 
\begin{equation*}
\mathbf{p}=[\mathbf{p}_{1},\mathbf{p}_{2},...,\mathbf{p}_{B_{2}}] \in R^{(L-K+s)\times B_{2} \times d}
\end{equation*}
Extensive computational resources are required when there is a large document as input. 
Hence, the capsule compression is conducted to condense the capsule number to a smaller one. 
The condensed capsule $u_{i}$ is computed as:
\begin{equation}
u_{i} = \sum_{j}w_{j}\mathbf{p}_{j} \in R^{d}
\end{equation}
where $w_{j}$ is the parameter needed to be learned.

In the next step, the transformation matrix $W_{ij}$ is utilized to generate the prediction vector $\hat{u}_{j|i}\in R^{d}$ (the parent capsule $j$) from its child capsule $u_{i}$, where $N$ is the number of parent capsules in the last step which is $(L-K+s)$.
This vector can be computed as follows:
\begin{equation}
\hat{u}_{j|i}=W_{ij}u_{i}+b_{j|i}\in R^{d}
\end{equation}where $b_{j|i}$ denotes the capsule bias term.
This step helps learn child-parent relationships for the capsule networks as argued in~\cite{sabour2017dynamic}.
Therefore, the feature map generated from the primary capsule layer is $\mathbf{u}\in R^{E\times N \times d}$, where $E$ is the capsule number after the matrix transformation.

Then the graph routing algorithm is utilized to get voting results, which can be depicted as:
\begin{equation}
v =\text{Graph-Routing}(\hat{u})
\end{equation}
where $\hat{u}$ is the composition of all of the child capsules, and $v\in R^{C\times d}$ denotes all of the parent-capsules, where $C$ denotes the class number. 
Details are referred to Section~\ref{sec:graph_routing}.
 
\subsection{Representation Layer}
Representation layer is the last layer which gets the input $v$ from the graph routing.
The final capsule number $C$ is the same as the class number of the document.
 The vector normalization is utilized to get the class probability $\hat{p}=\left\|v\right\| \in R^{C}$ .
 
\section{Graph Routing }
\label{sec:graph_routing}
 
\begin{figure}[ht]
 \centering
   \includegraphics[width=0.9\linewidth]{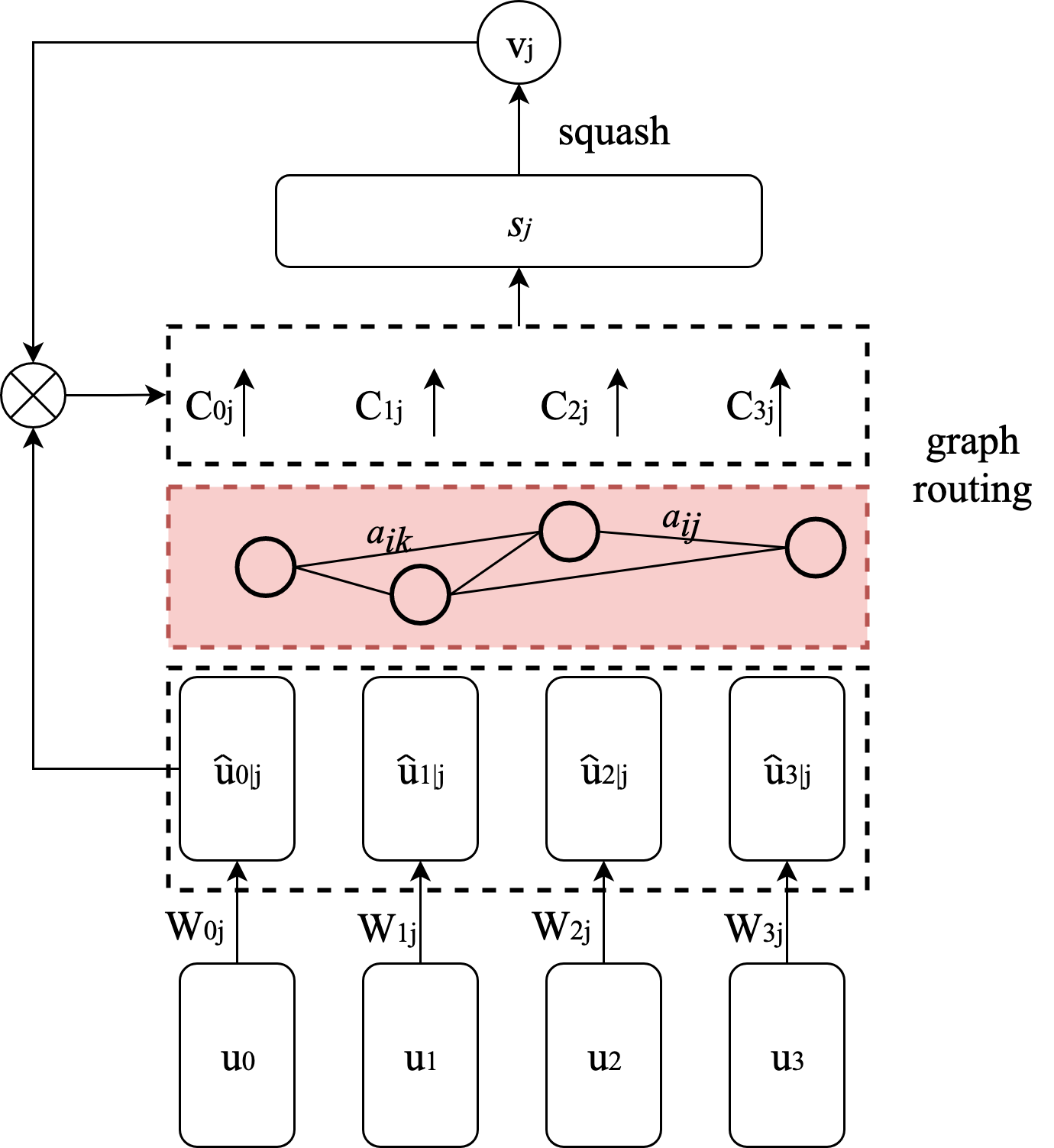} 
  \caption{The structure of the graph routing.}
  \label{fig:structure}
 \end{figure}

To make the routing simple and effective, each capsule is treated as a node in a graph.
Meanings of sentences are expressed by the composition of different phrases, 
and each capsule only captures a certain phrase during the feature selection. 
To obtain a comprehensive understanding of a sentence, 
the relationship between different capsules must be learned.
Actually, there should be connections between nodes as it will be helpful for reaching an agreement during the routing. 
Inspired by GCN~\cite{kipf2016semi}, which elaborates the relationships among the nodes in a graph, we learn intra-correlation with the graph convolutional operation. 
The structure of graph routing is shown in Figure~\ref{fig:structure}.

The GCN procedure is shown in the red box in Figure~\ref{fig:structure}, based on an undirected complete graph that is composed by the capsules in the lower layer.
There are usually more than three iterations in a routing. Therefore, in order to make the structure efficient and simple, a single GCN layer is used in each iteration to gather intra-relationship within the nodes.
If the number of iterations is set to 3, we will have a routing network with three layers of GCN.
The weight $a_{ij}$ in the edge is the intra-relationship between the two connected capsules $i$ and $j$.
Generally, if two connected capsules are semantically close, there will be a large value for the weight $a_{ij}$.
And if there is no semantic connection between these two capsules, there will be a small value for weight $a_{ij}$.
The way to get the intra-relationship between two capsules will be discussed in the following subsection.
In general, these three distance measurements are used first, coupled with normalization trick in subsection~\ref{subsec:Normtrick}, when evaluating the relationship between capsules. The attention mechanism in subsection~\ref{subsec:att} involved is the same as that in~\cite{bahdanau2014neural}.
 \subsection{Relationship Between Capsules}
 
In the common GCN~\cite{kipf2016semi}, the relationship between two nodes is expressed by the non-negative value in the adjacency matrix $A$. 
However, it is the prior known information for $A$ which is predefined by the existing data.
The convolutional operation is conducted in the Fourier domain to get the hidden feature, which is expressed in equation~\ref{equ:gcn}~\cite{hammond2011wavelets}.
\begin{equation}
\label{equ:gcn}
W_{i}\star u_{i} \approx W_{i} \tilde{A} u_{i} 
\end{equation}
where $W_{i}$ is the parameter of the network, and $\tilde{A}$ is from equation~\ref{equ:norm}
\begin{equation}
\label{equ:norm}
\tilde{A}=I_{N}+D^{-\frac{1}{2}}AD^{-\frac{1}{2}}
\end{equation}
where $A\in R^{N\times N}$ is the adjacency matrix, $N$ is the capsule number, and $D$ is the degree matrix, with $D_{ii}=\sum_{j} A_{ij}$.

After filtering by the convolutional kernel, $u_{i}$ will aggregate more relevant information for the classification.
 From equation~\ref{equ:gcn}, we can see that if one capsule is related to another capsule in layer $l$, the corresponding value in the adjacency matrix will be big.
The relationship matrix is one of the adjacency matrices, with the correlation value as the connection strength.
Generally, the adjacency matrix $A$ is known information where its values are binary or real to express the distance between two nodes. 

In our case, it is difficult to evaluate the relationship between two capsules as this information is unknown before routing.
Furthermore, the capsule vector is a hidden representation which lacks detailed semantic meaning.
Therefore, to overcome this challenge, we use different distance measurements to find the relationship among capsules semantically and effectively. 
In this paper, we explore three distance measurements, which are Wasserstein distance (WD), Euclidean distance (ED), and cosine similarity (CS).
We argue that if two vectors (for example, vector $y_{i}$ and vector $y_{j}$) express a similar meaning, they will be semantically close with a large relationship value in the adjacency matrix. 

 \subsubsection{Wasserstein Distance (WD)}
Wasserstein Distance is usually used in generative models~\cite{liigen} due to its ability in measuring the distance between probability distributions.
The capsule vector can be treated as the probability of a certain attribute that is captured.
Therefore, WD which is calculated as $d_{ij}^{w} = \inf E[\|y_{i}-y_{j}\|_{p}]^{1/p}$ can be applied in capsule vector relationship evaluation, where $\inf$ denotes infimum, i.e., the greatest lower bound. In this paper, $p$ is set to $1$. To transfer the Wasserstein distance $d^{w}$ to the intra-relationship, $a_{ij} = {-d^{w}_{ij}}$ is applied. 
Generally, if two capsules are semantically close, the value of the intra relationship will be big.

 \subsubsection{Euclidean Distance (ED)}
Euclidean Distance is the straight-line distance between two vectors in Euclidean space, and it is calculated as $d^{e}_{ij}=\sqrt{(y_{i}-y_{j})\cdot(y_{i}-y_{j})}$.
To transfer the Euclidean distance $d^{e}$ to the intra relationship, $a_{ij} ={-d^{e}_{ij}}$ is also applied.
Similarly, if two capsules are semantically close, the value of the intra relationship will be big.

 \subsubsection{Cosine Similarity (CS)}
Cosine Similarity is another popular measurement between two non-zero vectors, and it can be calculated as $d^{c}_{ij}=\frac{y_{i}\cdot y_{j}}{\left \|y_{i}\right \| \left\|y_{j}\right\|}$.
To ensure the distance to itself is zero which is same as the case in ED and WD, the intra relationship between two capsules is expressed as $a_{ij}=d^{c}_{ij}-1$.

 \subsection{Normalization Trick}
 \label{subsec:Normtrick}
Generally, the renormalization trick is used in equation~(\ref{equ:norm}), which is $I_{N}+D^{-\frac{1}{2}}AD^{-\frac{1}{2}}\rightarrow \hat{D}^{-\frac{1}{2}}M\hat{D}^{-\frac{1}{2}}$, with $M=A+I_{N}$ and $\hat{D}_{ii}=\sum_{j}M_{ij}$.
However, if we use ED and WD as the measurement for intra relationship, the values are negative, which makes it difficult to apply the general normalization trick to get the approximate adjacency matrix $A$ because it requires all of the values being non-negative.
To overcome this issue, a more general normalization trick for equation~\ref{equ:norm} is proposed in this paper, which is given in equation~\ref{equ:graph_a},
\begin{equation}
\label{equ:graph_a}
\tilde{A} \rightarrow \text{softmax}(A)+I_{N} 
\end{equation}
Each capsule treats itself as the centroid of the clustering by adding a one-hot vector.
$\tilde{A}$ has eigenvalues in the range $[0,1]$ as the distance to itself is $0$. 
And this avoids the gradients exploding/vanishing during the neural network optimization.
It is an instance-level normalization for the capsule's relationship learning.
Different from the renormalization trick mentioned before, the semantic distance is normalized in the view of the current capsule by using the softmax function in the same row.
It can be treated as a variant of 
random Laplacian matrix normalization.
This matches the true situation in which the importance of the word is different when the centroid word is different.
Taking ``The battery has a long life'' as an example, in the view of the word ``The'', the word ``battery'' will have a high value as ``The" is the definite article for ``battery''. 
However, in the view of the word ``battery'', ``long life'' will play a more important role to express the positive sentiment than the word ``The''. 
The ablation study shows the effectiveness of the proposed method.

\subsection{With Attention}
\label{subsec:att}
To further improve the quality of the learned features from our graph routing, and guarantee the learned features are useful, the attention mechanism is applied.
Graph routing is a bottom-up method which highly depends on lower layer features.
If there is no clear cluster center, it will be difficult to do the feature selection.
Different from the routing mechanism, attention mechanism is a top-down method which helps the routing to select 
useful relationships by considering the context information in the same layer. This guarantees the quality of the gathered relationship.
The attention is applied to the aggregated output $h_{j}$ from GCN which is described in equation~\ref{equ:atten}.

\begin{equation}
\label{equ:atten}
\begin{aligned}
&m_{j}\leftarrow f_{att}(h_{j})\\
&\alpha_{j} = \frac{\exp m_{j}}{\sum_{j} \exp m_{j}}
\end{aligned}
\end{equation}
Here, $f_{att}$ is the feed-forward function to get the attention weight. 
The attention score $\alpha_{j}$ is then applied to the learned feature $h_{j}$ that gets from GCN.
The proposed algorithm is in Algorithm~\ref{alg:1}. 
It is the same framework as dynamic routing, except for GCN and attention mechanisms.
As mentioned earlier, in a routing iteration, it is a single-layer GCN for the feature aggregation. 

\begin{algorithm}
\caption{The Algorithm of the Graph Routing}
\label{alg:1}
\begin{algorithmic}[1]
\REQUIRE Capsules $u_{i}$ 
\FOR{$l$ layers}
\STATE Initialize $e_{ij}\xleftarrow{} 0$ 
\STATE Get the adjacency matrix $A$ about capsules in layer $l$.
\STATE Get $\tilde{A}$ with equation~\ref{equ:graph_a}.
\STATE $h_{j} \leftarrow u_{ij}^{l}\tilde{A}_{i}W_{ij}$
\STATE $m_{j}\leftarrow f_{att}(h_{j})$
\STATE $\alpha_{j}=\text{softmax}(m_{j})$
\STATE $o_{j}\leftarrow \alpha_{j}h_{j}$
\FOR{$r$ routing number}
\STATE All capsule $i$ in layer $l$: 
\STATE\quad $c_{ij}\xleftarrow{}\text{softmax}(e_{ij}^{l})$
\STATE All capsule $j$ in layer $(l+1)$:
\STATE \quad $s_{j}\xleftarrow{}\sum_{i}c_{ij}o_{j}$ 
\STATE\quad $v_{j}\xleftarrow{} \text{squash}(s_{j})$
\STATE All capsule $i$ in layer $l$ and capsule $j$ in layer ($l$+1): 
\STATE \quad $e_{ij}^{l+1}\xleftarrow{}e_{ij}^{l}+u_{ij}^{l}\cdot v_{j}$ 
\ENDFOR
\RETURN $v_{j}$
\ENDFOR
\end{algorithmic}
\end{algorithm}
In the lines 3-4, the $\tilde{A}$ is obtained from the adjacency matrix $A$ row by row.
Lines 5-17 are the routing details about the graph routing.
In lines 5-8, GCN and attention are applied.
In the line 11, the weight scalar $c_{ij}$ is calculated from $e_{ij}$ in layer $l$.
In the line 14, the squash function is applied to get the output value $v_{j}$.
In the line 16, $e_{ij}$ in layer $l+1$ is calculated.

\section{Experiments}
\label{sec:experiments}
In this section, we conduct experiments to evaluate the effectiveness of the proposed graph routing. Specifically, we conduct experiments of document classification to validate the effectiveness of the proposed routing in text classification.
Then, we validate the hierarchical relationship learning with semantic consistency, and validate intra relationship learning with adjacency matrix illustration.
Next, the ablation study is conducted to validate the effectiveness of the proposed normalization trick. To show the generalization of the proposed routing, 
we also compare to BERT~\cite{devlin2018bert}.
Next, the case study is illustrated to see the effectiveness of the proposed routing.
Finally, we conduct a parameter analysis.
 
\subsection{Experimental Settings}

\subsubsection{Datasets}
We adopt five benchmark datasets for document classification to evaluate the effectiveness of the proposed framework. 
Amazon-Clothing and Amazon-Beauty are two review datasets that we select
\footnote{http://jmcauley.ucsd.edu/data/amazon/}, where each document is labeled with a rating ranging from $[1, 5]$.
Emotion dataset\footnote{https://data.world/crowdflower/sentiment-analysis-in-text} contains 13 emotion categories, namely: $\{$Love, Empty, Relief, Anger, Surprise, Neutral, Happiness, Sadness, Fun, Enthusiasm, worry, boredom, hate$\}$.
Yelp\footnote{http://yelp.com/dataset} is another review dataset we use.
The label range in Yelp dataset is also $[1,5]$.
AG News~\cite{conneau2016very} is a news dataset where each document belongs one of the four classes, i.e., world, sports, business, and technology .
The statistics of those datasets are listed in Table~\ref{tab:datasets}.
 
\begin{table*}[ht]
\centering
\begin{tabular}{l|c|c|c|c|c|c}
\hline
Datasets   & Train & Test  & Classes& Avg-Docs & Max-Docs & Vocab Size\\ \hline
Amazon Clothing & 250k  &25k &5  & 68.0 & 4355 & 81836  \\ \hline
Amazon Beauty    &180k &15k &5 &99.2 & 4494 &80935 \\ \hline 
Emotion &15k &5k &13 &65.7 & 167 & 30716 \\\hline
Yelp       &  650k & 50k  & 5 & 118.4 &1463 & 293402 \\  \hline 
AG news  & 120k & 7.6k & 4  &  39.3 &196& 19637   \\ \hline
\end{tabular}
\caption{The details of the datasets.}
\label{tab:datasets}
\end{table*}

Avg-Docs denotes the average word number in the document, while Max-Docs is the maximum word number. Vocab size is the number of the word that the corpus contains.

\subsubsection{Implementation Details}
We use TensorFlow in python to implement our model.
In all of the experiments, all word embeddings are initialized randomly with 300-dimensional vectors.
The capsule number is set to 50, the out channel number from NCL and PCL are 64, the stride for the convolutional operation is 2. 
The batch size is set to 32, and the learning rate is 5e-5 during the model training with the Adam optimization algorithm~\cite{kingma2014adam}. 
All of the experimental results are averaged over 5 runs. 
 
\subsection{Baseline Methods}
In the experiments, the compared routing methods include:
\begin{itemize}
\item \textbf{EM Routing:} EM routing is proposed in~\cite{sabour2018matrix}. It is good at spatial feature extraction, and usually used in image classification.
\item \textbf{Dynamic Routing:} Dynamic routing is another popular routing method which is proposed in~\cite{sabour2017dynamic}, and also usually used in image classification.
\item \textbf{Leaky Dynamic Routing:} Leaky dynamic routing is proposed in~\cite{zhao2018investigating} and achieves state of the art results in sentiment classification.
\item \textbf{KDE Routing:} KDE routing is another robust routing model proposed in~\cite{zhao2019towards}. This model is good at the multi-label classification and has a good generalization.
\item \textbf{GCN:} To make the ablation study about the GCN, we only utilize GCN in the routing part. Here, the adjacency matrix is calculated with WD.
\item \textbf{Transformer:} The transformer-based method are very 
strong baseline models. 
It is added on top of the model to replace the routing part while keeping other parts identical. 
\end{itemize}

To make a fair comparison, all of the compared methods share the same parameters with the same architecture described in section~\ref{sec:architecture}.

\subsection{Document Classification}

The results with different routing methods are listed in Table~\ref{tab:res1}.
The bold number in the table is the best result in a dataset.

\begin{table*}[ht]
\centering
\begin{tabular}{p{3.3cm}|p{1.3cm}|p{1.3cm}|p{1.3cm}|p{1.3cm}|p{1.3cm}}
\hline
Routing Methods        & Amazon Clothing & Amazon Beauty & Microsoft Emotion & Yelp          & AG News \\ \hline 
\textbf{Transformer}         & 61.87\%  & 62.50\%  & 31.07\% &64.61\% & 89.16\%   \\\hline
\textbf{EM Routing}         & 60.57\% & 60.65\% & 31.04\% & 62.51\% & -\\
\textbf{Dynamic Routing}    & 62.87\%  & 63.19\%  & 30.82\% & 63.65\% & 89.02\%    \\ 
\textbf{Leaky DR (A)} & 62.92\% &63.47\%  & 31.11\% & 64.08\% & 92.10\%$\S$   \\ 
\textbf{Leaky DR (B)} & - & - & - & - & 92.60\%$\S$   \\ 
\textbf{KDE Routing}          & 62.94\% &63.95\%   &31.83\% &64.45\% & 90.08\%\\\hline
\textbf{GCN}                  &62.86\% & 63.47\%  &30.87\%   & 63.35\% &89.61\% \\\hline
\textbf{Graph Routing(CS)}     &62.80\% & 63.81\%  &31.40\%   & 64.69\% & 92.30\%\\
\textbf{Graph Routing(ED)}    &63.19\% & 64.03\%   &31.55\%   & 64.61\% & 91.91\% \\
\textbf{Graph Routing(WD)}    &63.44\% & 64.13\%   & 31.61\%  & 65.33\%  & \textbf{92.62\% }\\\hline
\textbf{With Attention}         &\textbf{63.76\%}& \textbf{64.34\%} & \textbf{31.90\%} & \textbf{65.62\%} & 92.60\%\\\hline
\end{tabular}
\caption{The classification results with different routing methods over those five datasets. $\S$ means results citing from ~\cite{zhao2018investigating}. Leaky DR is the abbreviation of leaky dynamic routing that proposed in ~\cite{zhao2018investigating}. }
\label{tab:res1}
\end{table*}
The word in the bracket behind the graph routing represents the intra relationship measurement used in the adjacency matrix.
In the table, we can see that our proposed graph routing achieves competitive performance.
In particular, graph routing with different intra relationships substantially outperforms the transformer, and there is a noticeable margin on all the experimental datasets.
Also, graph routing with different intra relationships obtain competitive results against existing routing methods, such as dynamic routing, leaky dynamic routing and EM routing, 
especially in the graph routing with WD which achieves the best results on all of the datasets compared with other intra relationship measurements.
Therefore, we can conclude that the graph routing with WD has obtained a good quality of the relationship between different capsules.
There is a slight improvement for the leaky dynamic routing by using Leaky-Softmax~\cite{sabour2017dynamic} to replace the standard softmax function in strengthening the relationship between a child and parent capsules. 
If we only apply GCN to map the capsule vector to the final output by aggregating the learned features directly, the improvement compared with existing routing methods is limited. 
By combining with the attention mechanism, our graph routing with WD has a further improvement with the best precision on most of the datasets compared with other routing methods. 
That is to say, with the help of the attention mechanism, graph routing becomes more robust and effective.

 \subsection{Semantic Consistency }

One important feature of our graph routing method is that it can learn the hierarchical and intra relationship between capsules at the same time. In this subsection, the effectiveness in the hierarchical relationship learning is validated.
However, it is difficult to evaluate this relationship directly as there is no specific meaning in the capsule, even after the graph routing.
However, if this relationship between different capsules are well learned in a layer, the capsule distribution in that layer will correlate with the final output as well.
That is to say, there will be a high semantic consistency between the current capsule layer and the final output if the hierarchical relationship is well learned.
Thus, semantic consistency between different layers is calculated.
The final output denotes $v\in R^{C\times d}$, where $C$ is the class number, the output in each layer is transferred to $p\in R^{m\times d}$.
Each vector in $p$ will be labeled with the class number by finding the closest vector in $v$. 
Then semantic consistency is calculated as the percentage of times that the vector is in the right class. 
And the vectors from the NCL, PCL, and RL are considered.
Then the outputs from those three layers will compute with the final output to get the semantic consistency.
This experiment is conducted on the AG News dataset, and the results are listed in Table~\ref{tab:relationship}.

\begin{table}[ht]
\centering
\begin{tabular}{l|l|l|l}
\hline
Routing Methods       & NCL    & PCL    & RL   \\\hline
Dynamic Routing    & 25.72\% & 25.39\% & 85.82\% \\\hline
Leaky Dynamic Routing & 21.82\% & 25.05\% & 85.39\% \\\hline
KDE Routing      & 24.65\% & 25.56\% & 86.30\% \\\hline
Graph Routing     & 24.85\% & 25.70\% & 87.55\% \\\hline
\end{tabular}
\caption{The semantic consistency between different layer output.}
\label{tab:relationship}
\end{table}

From the Table~\ref{tab:relationship}, we can see that there is no difference in column NCL and PCL compared with existing routing methods. 
There are four categories about the document in the AG News.
Therefore, only about 25\% of the vectors from layer NCL and PCL remain relevant to the final output.
However, after the routing, the number that related to the final output increases sharply, especially in the case of graph routing.
That is to say that the hierarchical relationship between different capsules is learned well after the graph routing.

\subsubsection{Different Relationship Measurements}
\begin{figure}[ht]
 \centering
 \includegraphics[width=0.9\linewidth]{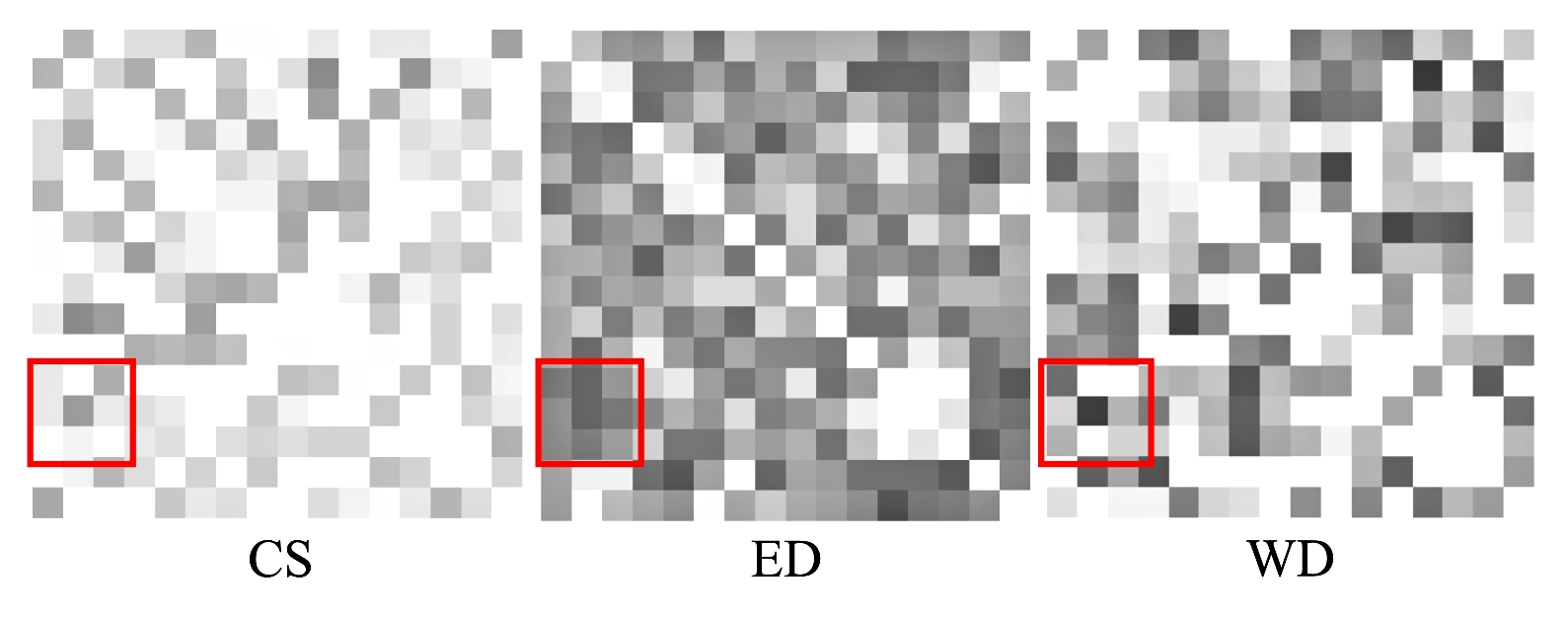}
    \caption{The relationship among capsules that learned with different measurements.}
  \label{fig:sketch}
 \end{figure}

There are different intra relationship measurements to get the adjacency matrix between the capsules. 
From Table~\ref{tab:res1}, we can see that the graph routing with WD achieves the best results on all datasets compared with CS and ED. 
That is to say that WD is an efficient method in distance evaluation.
To explain the goodness of the WD, the adjacency matrices with different measurements over the same document are illustrated in Figure~\ref{fig:sketch}.

 From the color blocks in the red box which is selected randomly from the same location in the three cases, we can see that the ED are noisier than the other two measurements, and there is a clear value in WD. 
 On the contrary, the value in CS is blurred. 
 That is to say that WD has a good evaluation for the intra relationship.

\subsection{Ablation Study}
\begin{figure}[H]
 \centering
 \includegraphics[width=0.8\linewidth]{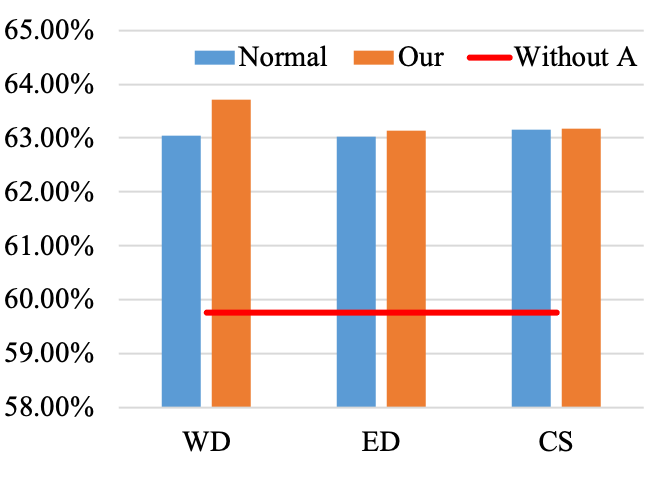}
    \caption{The classification results with different normalization trick in GCN. $y$-axis denotes the classification result, and the red line named `Without A' is the result of $\tilde{A}$ being replaced by the identity matrix.}
  \label{fig:ablation1}
 \end{figure}

To analyze the effectiveness of the new normalization trick we proposed in subsection~\ref{subsec:Normtrick}, the ablation validation compared with the renormalization in equation~\ref{equ:norm} is conducted.
As there is no negative value for the adjacency matrix, the exponential function is applied to the ED and WD.
Therefore, $a_{ij}=\exp(-d^{e}_{ij})$ in the case of ED, and $a_{ij}=\exp(-d^{w}_{ij})$ in the case of WD.
CS is kept the same due to its range being $[0,1]$.
Except for the normalization trick over $\tilde{A}$ is different, all the other parts of the model are kept the same.
The results are showed in Figure~\ref{fig:ablation1}.

From the figure, we can see that our normalization trick can achieve better results compared with the normal one, especially in the case of WD and ED.
When using CS as the relationship measurement, there is no difference between those two normalization tricks as our normalization trick can be treated as a variant of random Laplacian matrix normalization.

To validate the importance of the intra-relations that learned by the graph routing, we replace $\tilde{A}$ that learned from Equation~\ref{equ:norm} with the identity matrix, and mask the intra-relations that learned among the capsules in the same layer. The result is pictured in the red line named `Without A' in Figure~\ref{fig:ablation1}. From the red line we can see that, all these three different distance metrics are not working under such condition. That is to say, without intra-relations, the performance of our graph routing degrades dramatically. This validates the importance of the intra-relations that learned by our graph routing. 

 \subsubsection{With BERT}
BERT~\cite{devlin2018bert} is so popular that we can not ignore it. To make fair comparisons, we set ``output all encoded layers'' to \textit{True} in the pre-trained BERT~\footnote{https://github.com/google-research/bert} model, which has been trained on massive data and computing resources. Before sending these results (12 layers in total, we also select last 5 layers as another example) to the classifier, there are four ways to produce the output, which are max pooling out (Max), average pooling out (Avg), summation (Sum) and our routing method. The results on the Microsoft emotion dataset are reported in Table~\ref{tab:with_bert}.
\begin{table}[ht]
\centering
\begin{tabular}{l|l|l|l|l}
\hline
          & Max      & Avg       & Sum  & Our Routing\\\hline
5 layers   & 36.13\%  &  37.26\%   &33.08\% & \textbf{37.41\%}   \\\hline
12 layers  &36.43\%   &  37.00\%   & 31.77\% &\textbf{37.03\%} \\\hline  
\end{tabular}
\caption{Comparisons with different pooling out methods in BERT.}
\label{tab:with_bert}
\end{table}
 
We can see that with the help of BERT, our graph routing achieves the best results compared with other pooling methods both in the case of 5 layers and 12 layers. From this results, we can conclude that our graph routing has a good generalization when applied in a new model.

\subsection{ Case Study}
\begin{figure}[ht]
 \centering
 \includegraphics[width=1\linewidth]{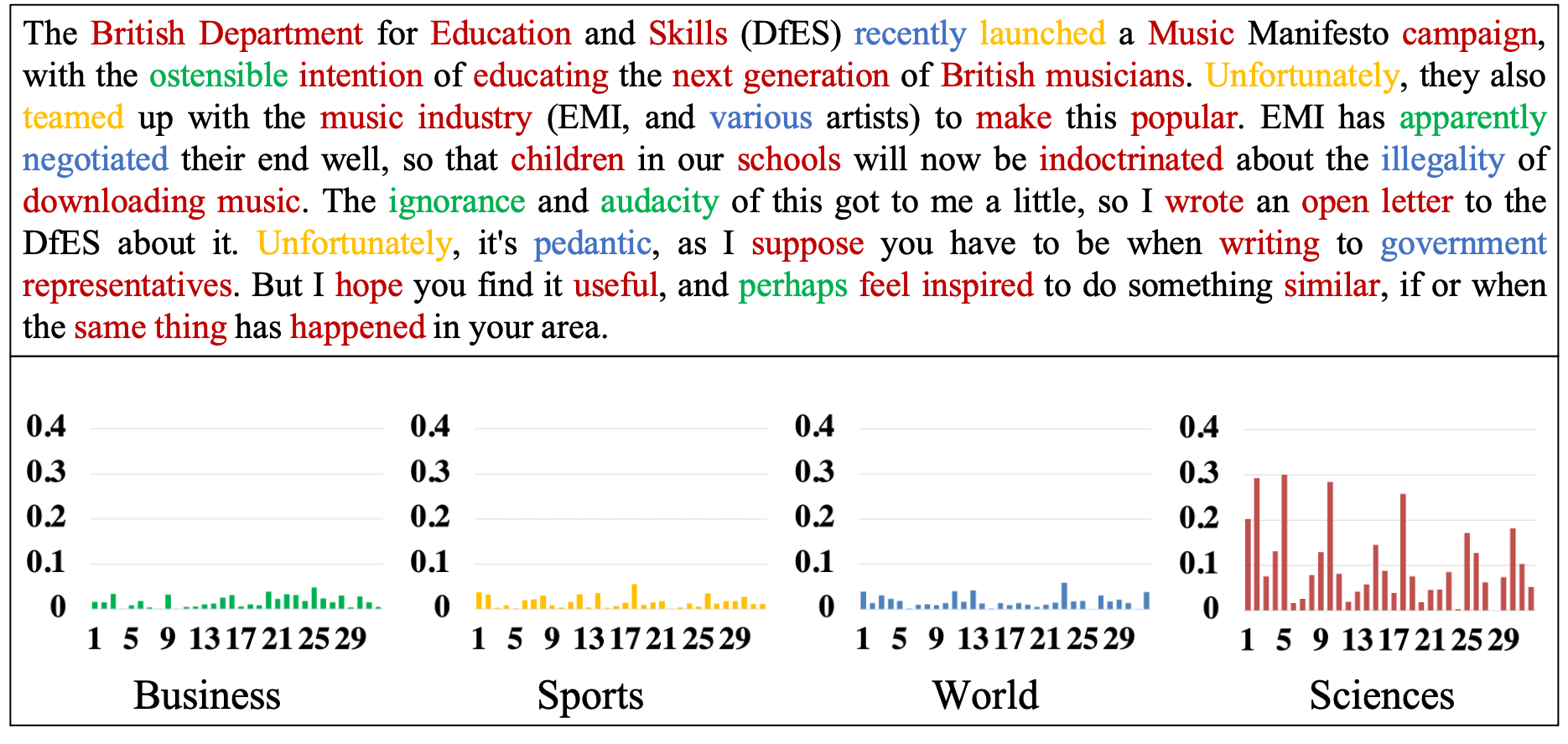}
    \caption{The figure above is the distribution of the word. Words in red are the science category, in blue are the world, in yellow are the sport and in green are the business. The figures below are the category vector distribution after the squash operation.}
  \label{fig:casestudy}
 \end{figure}

In this subsection, we show in Figure~\ref{fig:casestudy} 
the effectiveness of our model in category information extraction.
Each word embedding is labeled by the vector from the graph routing. 
 
 We can see that most of the words are labeled as the science category after the optimization, and those words are closely related with the sentence topic. Also, from the output vector that is shown in the text below, we can see that the values in science are bigger compared with other categories. This example shows the effectiveness of our routing in the semantic relationships extraction.

\subsection{Parameter Analysis}
 \begin{figure}[ht]
 \centering
 \begin{subfigure}{.4\textwidth}
 \centering
 \includegraphics[width=0.98\linewidth]{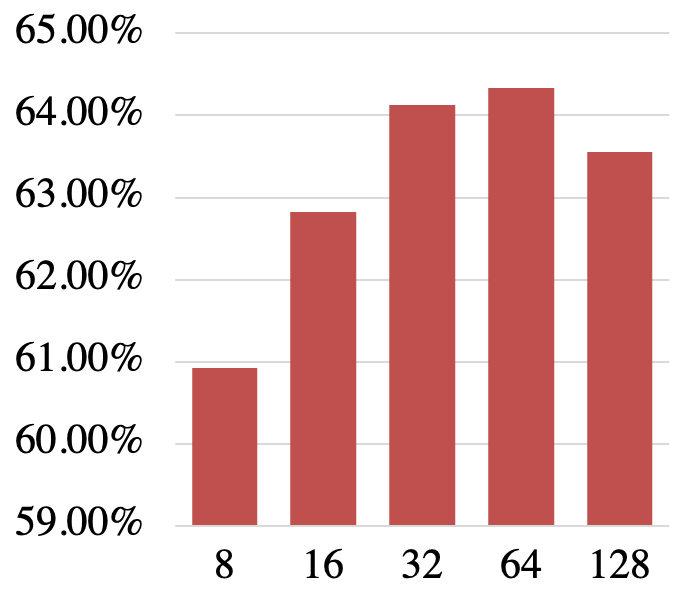}
   \caption{Capsule Number \label{subfig:capnum}}
 \end{subfigure} 
 \begin{subfigure}{.4\textwidth}
 \centering
 \includegraphics[width=0.98\linewidth]{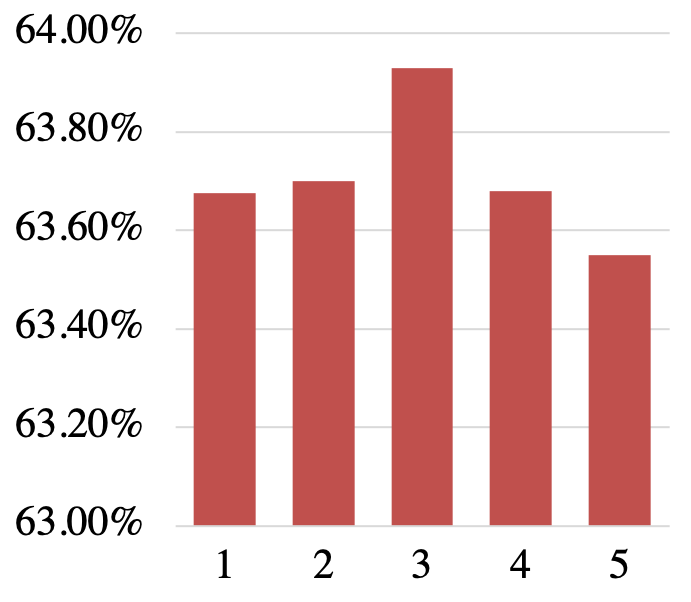}
   \caption{Iteration Number \label{subfig:iternum}}
 \end{subfigure}
   \caption{The parameters analysis about the capsule number. $y$-axis denotes the classification result.}
   \label{fig:parameters}
 \end{figure}
In this subsection, two parameters of the routing, iteration number and the capsule dimension, are discussed. 
Results are in Figure~\ref{fig:parameters}.
In Figure~\ref{subfig:capnum}, when the capsule dimension is set to 64, it achieves the best result. 
When the capsule dimension is 256, the server shows the OOM error due to the huge amount of memory occupation.
 There is no improvement compared with the dimension of 64 when the dimension is 128. Therefore, it is the best choice to set the dimension to 64.
In Figure~\ref{subfig:iternum}, when the iteration number is set to 3, it achieves the best result.
As the iteration number increases, there is no improvement in performance. 

 \begin{table}[htp]
 \centering
\begin{tabular}{l|l|l}
\hline
Routing methods   & \#Parameter & Time  \\\hline
EM Routing       & 31.69MB   & 0.1587$s$ \\ \hline
KDE Routing      & 31.68MB   & 0.1439$s$ \\\hline
Leaky DR (A)     & 69.07MB & 0.2634$s$ \\ \hline
Dynamic Routing  & 69.07MB & 0.2632$s$ \\\hline
Graph Routing with Attention      & 40.09MB & 0.2655$s$ \\\hline
\end{tabular}
\caption{The parameter number and time cost of different routing methods. $s$ denotes second.}
\label{tab:cost}
\end{table}
We know that graph routing has more steps than existing routing methods, for example, it takes time to compute the adjacency matrix with WD. To validate its computing efficiency,
a comparison of the number of parameters and the time cost is shown in the table~\ref{tab:cost}. 
From the table, we can see that our graph routing has a smaller parameter number than dynamic routing and leaky dynamic routing. Although the time cost is slightly larger than dynamic routing and leaky dynamic routing, it is still competitive considering the number of parameters.
 
\section{Related Works }
In this work, we applied the GCN in the routing between the capsules. 
Therefore, the related works about the routing between capsules will be introduced.
Also, our work relates with document classification with capsule network which will be introduced at the same time.
 
\subsection{Routing Between Capsules}
Since the concept of ``capsule'' neural networks~\cite{hinton2011transforming} was firstly proposed, it has become a hot research topic due to its ability in improving the representational limitation of CNNs in which pooling operation causes the information losing.
The transformation matrices between different capsules make the capsule network capture the part-whole relationships.
Together with the routing-by-agreement~\cite{sabour2017dynamic,sabour2018matrix}, capsules neural network has achieved promising results on MNIST data.

It is a bottom-up way for the dynamic routing and the EM routing proposed in~\cite{sabour2017dynamic,sabour2018matrix} when clustering the vector or matrix together.
There are different variants of dynamic routing.
Leaky softmax is applied by replacing the softmax function to strengthen the relationship between a child and parent capsules~\cite{zhao2018investigating}. 
Task routing algorithm was proposed by applying the clustering procedure in the task level~\cite{xiao2018mcapsnet}. 
Different from the general dynamic routing, task routing introduced a new coupling coefficient $c_{ij}^{k}$ by subdividing it to the task $k$. 
\cite{xi2017capsule} gave empirical results over the best parameters selection in the capsule network with dynamic routing. 
Especially the routing iterations during the routing procedure affects the performance.
\cite{chen2018generalized} believed that dynamic routing is not well integrated into the training procedure, especially for the iteration number which needs to be decided manually. 
Thus, there is a bottleneck for the dynamic routing due to its expense of computation during the routing.
Therefore, Zhang et al.~\citep{zhang2018fast} proposed weighted Kernel Density Estimation (KDE) to accelerate the routing.
In~\cite{zhao2019towards}, an adaptive KDE routing algorithm was proposed to make the routing decide the iteration number automatically, which gives a further optimization for the routing.
The empirical results demonstrate its effectiveness on different classification tasks.
Apart from KDE, K-means clustering method was also validated for its effectiveness during routing~\cite{ren2018compositional}.
It is a bottom-up way for the attention, while routing is a top-down way.
There are also researches about 
combining attention with the routing together in the capsule network~\cite{choi2019attention}.
 
 \subsection{Capsule Network For Text Classification}

Since capsule networks were proposed, most 
works are on image classification~\cite{sabour2017dynamic,sabour2018matrix,xi2017capsule,chen2018generalized}, 
while recurrent neural networks (RNNs) 
are more popular in text data processing~\cite{yang2018phd,liibie,cheens,chalea}.
In~\cite{wang2018sentiment,gong2018information}, RNN-Capsule was proposed for sentiment analysis. In~\cite{ren2018compositional}, together with the proposed compositional coding mechanism, bidirectional Gated Recurrent Units (GRU) were adopted 
for text classification. Based on the RNN, the attention model was applied in the capsule network 
for aspect-level sentiment analysis~\cite{wang2019aspect}.

Sabour et al.~\cite{sabour2017dynamic} applied the convolutional layer, primary capsule layer in their model similarly as in image classification. MCapsNet stacked the convolutional network, primary capsule layer, and representation layer together to do multi-task text classification~\cite{xiao2018mcapsnet}.
Different from the MCapsNet, Zhao et al.~\cite{zhao2018investigating} paralleled three such architecture together with filter window of 3, 4, 5 in the convolutional layer. 
All of those works show the effectiveness of such architectures in text feature extraction. 
 
\section{Conclusion \& Future Works }
It is the first time to treat the capsule as a node in a graph, and our graph routing applies GCN to explore the relationship between capsules in the same layer.
Together with the attention mechanism, graph routing also remedies the disadvantage of the routing which highly depends on the lower level features.
Furthermore, different distances between capsules are discussed.
Empirical results show the effectiveness of the proposed model: 
The performance of three out of five datasets is improved by more than 0.3 compared with state-of-the-art models. 
In the future works, we will explore a more effective graph neural network in the intra-relationship learning within the capsules, and find out a more efficient method for feature learning of text data using capsule networks. Also we will explore and design a new attention mechanism to accommodate our graph routing.
 \section{Acknowledgment}
This research is supported by the Agency for Science, Technology and Research (A*STAR) under its AME Programmatic Funding Scheme (Project \#A18A2b0046).

\bibliographystyle{plain}
\bibliography{references} 
\end{document}